\relax
\documentclass[11pt]{article}
\usepackage{fullpage}

\usepackage{amsthm}    
\usepackage{enumitem}  
\usepackage{siunitx}   

\usepackage[round]{natbib}
\usepackage[colorlinks=true,linkcolor=blue,citecolor=blue]{hyperref}
\usepackage{url}

\usepackage{graphicx}
\graphicspath{{./figures/}}
\usepackage{subcaption}

\usepackage{booktabs}
\usepackage{threeparttable}
\usepackage{multirow}

\usepackage{math_abbr}
\usepackage{math_op}
\usepackage{plate_notation}

\newtheorem{theorem}{Theorem}
\theoremstyle{definition}
\newtheorem{example}{Example}

\begin{document}
\title{Learning from Indirect Observations}
\author{Yivan Zhang$^{1}$ \and Nontawat Charoenphakdee$^{1,2}$ \and Masashi Sugiyama$^{2,1}$}
\date{
$^1$ The University of Tokyo 
$^2$ RIKEN AIP
}
\maketitle
\begin{abstract}
Weakly-supervised learning is a paradigm for alleviating the scarcity of labeled data by leveraging lower-quality but larger-scale supervision signals.
While existing work mainly focuses on utilizing a certain type of weak supervision, we present a probabilistic framework, \emph{learning from indirect observations}, for learning from a wide range of weak supervision in real-world problems, e.g., noisy labels, complementary labels and coarse-grained labels.
We propose a general method based on the \emph{maximum likelihood} principle, which has desirable theoretical properties and can be straightforwardly implemented for deep neural networks.
Concretely, a discriminative model for the true target is used for modeling the \emph{indirect observation}, which is a random variable entirely depending on the true target stochastically or deterministically.
Then, maximizing the likelihood given indirect observations leads to an estimator of the true target implicitly.
Comprehensive experiments for two novel problem settings --- \emph{learning from multiclass label proportions} and \emph{learning from coarse-grained labels}, illustrate practical usefulness of our method and demonstrate how to integrate various sources of weak supervision.

\end{abstract}

\section{INTRODUCTION}
Recent machine learning techniques such as deep neural networks mitigated the need for hand-engineered features, but still usually require massive hand-labeled training data from human experts \citep{lecun2015deep, goodfellow2016deep}.
In the real world, it is often infeasible to collect a large amount of labeled data due to high labeling costs, lack of domain expertise, or privacy concern \citep{horvitz2015data, jordan2015machine}.
The scarcity of high-quality hand-labeled data has become the bottleneck of further deployment of machine learning in the real-world environment.
Among other approaches addressing this labeled data scarcity problem, such as \emph{semi-supervised learning} \citep{olivier2006semi}, \emph{active learning}  \citep{settles2012active} and \emph{transfer learning} \citep{pan2009survey}, \emph{weakly-supervised learning} \citep{zhou2017brief} is a learning paradigm to leverage lower-quality but larger-scale supervision signals, which are cheaper and easier to obtain.

An example of weakly-supervised learning is \emph{learning from noisy labels} \citep{angluin1988learning, scott2013classification, natarajan2013learning, patrini2017making}, where we use high-quantity but low-quality labels provided by non-expert human annotators or web scrapers.
Another example for binary classification tasks is \emph{learning from positive and unlabeled data}, a.k.a. \emph{PU learning} \citep{elkan2008learning, du2014analysis}, where only positive and unlabeled data are given because negative data is difficult or impossible to collect, e.g., in land-cover classification \citep{li2010positive} or bioinformatics \citep{ren2015positive}.
For multiclass classification tasks, it might be easier for annotators to provide information about classes that an instance does \emph{not} belong to. 
This problem is called \emph{learning from complementary labels} and has been studied recently \citep{ishida2017learning, ishida2019complementary, yu2018learning}.

Among previous studies, one of increasingly popular approaches is to modify the objective function, i.e., \emph{loss correction} \citep{natarajan2013learning, van2017theory, lu2018minimal}.
In particular, \emph{forward correction} \citep{sukhbaatar2014training, patrini2017making} is a loss correction method to learn a classifier from noisy labels effectively.
Concretely, the noise transition matrix is multiplied after applying a softmax function to a deep neural network.
Then, noisy labels are compared with ``noisified'' predictions \citep{patrini2017making}.
For learning from complementary labels, \citet{yu2018learning} also proposed a similar loss correction technique.


\paragraph{Our Contribution} 
In this paper, we take a closer look at the forward correction method and point out that aforementioned methods used pervasively in different scenarios \citep{sukhbaatar2014training, patrini2017making, yu2018learning} are essentially the same method based on the \emph{maximum likelihood} principle, and can be used for learning from a wide range of weak supervision in real-world problems.
We show this fact by introducing a probabilistic framework called \emph{learning from indirect observations} (Section~\ref{sec:problem}).
The \emph{indirect observation} is defined as a random variable that only depends on the \emph{true target} (\emph{direct observation}) (see Figure~\ref{fig:xzy}).
The cardinality of the true target and the indirect observation can be different, which allows high flexibility to represent a broad class of weakly-supervised learning problems.
Next, we propose a method based on the \emph{maximum likelihood} principle to handle this problem (Section~\ref{sec:method}).
Concretely, a discriminative model for the true target is used for modeling the indirect observation.
Then, maximizing the likelihood given indirect observations leads to an estimator of the true target implicitly.
We can apply this method to other settings as long as they can be formulated as learning from indirect observations.
Moreover, we can naturally combine different types of indirect observations without having additional hyperparameters.

We also conduct theoretical analyses in Section~\ref{sec:analyses} by characterizing the behavior of our maximum likelihood estimator.
It is well-known that given \emph{direct observations}, the maximum likelihood estimator is \emph{consistent} under mild conditions \citep{lehmann2006theory}.
Here, we clarify conditions that are required for our estimator based on \emph{indirect observations} to be consistent (Section~\ref{ssec:consistency}).
We show that the only additional condition for the consistency is the identifiability of parameters.
Further, we propose to use the \emph{asymptotic variance} to measure how much information can be obtained from a certain type of indirect observation (Section~\ref{ssec:asymptotic_variance}).
Our analysis suggests that the asymptotic variance given some type of indirect observation could be large, thus more data are required compared with other type of indirect observation or direct observation.
This analysis can be used as a tool to balance the trade-off between the quality of labels and costs of the label collection process in real-world tasks.

Finally, to show practical usefulness of our framework, we conduct experiments in Section~\ref{sec:experiments} for two novel problem settings --- \emph{learning from multiclass label proportions} (Section~\ref{ssec:label_proportions}) and \emph{learning from coarse-grained labels} (Section~\ref{ssec:coarse_grained_labels}).
In experiments, we discuss the behavior of our model when assumptions on the data generating process are slightly violated, and demonstrate how to integrate various sources of weak supervision, e.g., coarse-grained labels and complementary labels.

\section{PROBLEM}
\label{sec:problem}
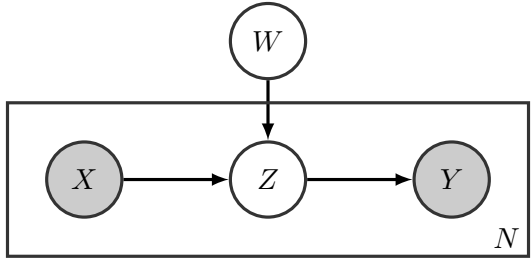
\begin{figure}[t]
\begin{center}
\begin{tikzpicture}
\platenotation
\node (X) [observable] {$X$};
\node (Z) [unobservable, right = of X] {$Z$};
\node (Y) [observable, right = of Z] {$Y$};
\path (X) edge [dependency] (Z);
\path (Z) edge [dependency] (Y);
\node (W) [unobservable, above=8mm of Z] {$W$};
\path (W) edge [dependency] (Z);
\node [plate, fit=(X)(Z)(Y), inner sep=5mm,
       label={[anchor=south east] south east: $N$}] {};
\end{tikzpicture}
\end{center}
\caption{\textbf{Graphical representation of the data generating process.} 
Here, $X$ and $Y$ represent the \emph{feature vector} with the \emph{indirect observation}, while $Z$ is the unobservable \emph{true target}.
$Z$ follows a parametric distribution $p(Z \vert \theta)$ parameterized by $\theta = f(X; W)$, which is the output of a deterministic function of $X$, parameterized by $W$.
The goal is to estimate $W$ from observation of $(X, Y)$-pairs so that we can predict the true target $Z$ from the feature vector $X$.}
\label{fig:xzy}
\end{figure}


Consider a prediction problem, such as classification or regression.
Let $X \in \sX$ and $Z \in \sZ$ be random variables representing the \emph{feature vector} and the \emph{true target} (\emph{direct observation}), respectively,
where $\sX$ and $\sZ$ denote their support spaces.
The \emph{indirect observation} $Y \in \sY$ is a random variable that entirely depends on a single instance of true target $Z$,
according to a conditional probability $p(Y \vert Z)$. 
In learning from indirect observations, we are given an i.i.d.~sample of $(X,Y)$-pairs 
$\{x_i, y_i\}_{i=1}^n \iid p(X, Y)$\footnote{
In this work, uppercase letters $X, Y, Z$ are random variables, and lowercase letters $x, y, z$ are instances of random variables.
Abusing notation, $p(\cdot)$ denotes a distribution and also its probability mass/density function.
}.
The goal is to learn a model that predicts the true target $Z$ from the feature vector $X$.
Note that the true target $Z$ is not observed.
Figure~\ref{fig:xzy} illustrates the graphical representation of the data generating process.


\begin{table*}[t]
\caption{\textbf{Examples of Learning from Indirect Observations}}
\label{tab:examples}
\begin{center}
\begin{threeparttable}
\begin{tabular}{llll}
\toprule
Cardinality 
& Learning from ... 
& True Target $Z$ 
& Indirect Observation $Y$ 
\\ \midrule

\multirow{4}{*}{$\abs{Z} = \abs{Y}$} 
& positive and unlabeled data\tnote{1}
& \{positive, negative\} 
& \{positive, unlabeled\} 
\\ \cmidrule{2-4}

& noisy labels\tnote{2}
& which class $X$ belongs to
& which class $X$ \emph{might} belong to
\\ \cmidrule{2-4} 

& complementary labels\tnote{3}
& which class $X$ belongs to
& which class $X$ does \emph{not} belong to
\\ \midrule

$\abs{Z} \leq \abs{Y}$ 
& \begin{tabular}[c]{@{}l@{}}
multiclass label proportions\tnote{*} \\
(Section~\ref{ssec:label_proportions})
\end{tabular} 
& which class $X$ belongs to
& which \emph{group} $X$ belongs to
\\ \midrule

$\abs{Z} > \abs{Y}$ 
& \begin{tabular}[c]{@{}l@{}}
coarse-grained labels\tnote{*} \\
(Section~\ref{ssec:coarse_grained_labels})
\end{tabular}
& which class $X$ belongs to
& which \emph{super-class} $X$ belongs to
\\ \bottomrule
\end{tabular}
\begin{tablenotes}
\footnotesize
\item[1] a.k.a.~PU learning, in the censoring setting \citep{elkan2008learning}.
Another subtly different setting is the case-control setting \citep{ward2009presence, du2014analysis}.
A comparison can be found in Appendix~\ref{app:pu}.
\item[2] Class-conditional noise (CCN) \citep{angluin1988learning, natarajan2013learning, patrini2017making}.
\item[3] Uniform \citep{ishida2017learning} or biased complementary labels \citep{yu2018learning}.
\item[*] Proposed problem settings. See corresponding sections for details.
\end{tablenotes}
\end{threeparttable}
\end{center}
\end{table*}

\begin{figure*}[t]
\begin{center}
\includegraphics[width=.85\linewidth]
{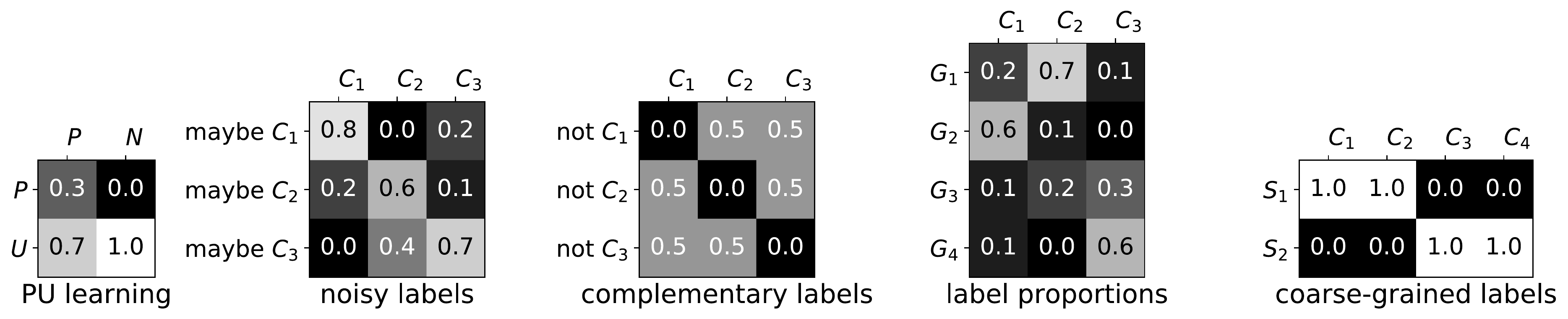}
\caption{\textbf{Examples of conditional probability $P(Y \vert Z)$ in the matrix form}.
Columns represent the true target $Z$, rows represent the indirect observation $Y$. 
Each column sums to $1$.}
\label{fig:conditional_probability}
\end{center}
\end{figure*}


Concretely, we assume that the joint distribution $p(X, Z, Y)$ can be factorized as follows:
\begin{equation}
\label{eq:factorization}
\begin{aligned}
  p(X, Z, Y)
&= 
  p(Y \vert X, Z) p(Z \vert X) p(X)
\\
&=
  p(Y \vert Z) p(Z \vert X) p(X)
,
\end{aligned}
\end{equation}
i.e., we assume $p(Y \vert X, Z) = p(Y \vert Z)$.
This means that $Y$ entirely depends on $Z$, not on $X$.
This restriction is used pervasively in previous studies \citep{elkan2008learning, patrini2017making, yu2018learning}.
However, in real-world problems, this restriction could be violated to some extent.
We explore such situations experimentally in Section~\ref{ssec:label_proportions}.
Several learning problems that can be formulated as learning from indirect observations are provided in Table~\ref{tab:examples}.

The conditional probability $p(Y \vert Z)$ is crucial for learning from indirect observations.
It can be estimated from data, observed, or determined by the type of indirect observation.
For example, for learning from noisy labels, \citet{patrini2017making} proposed a method to estimate the conditional probability $p(Y \vert Z)$, called the \emph{noise transition matrix} in this scenario;
for learning from complementary labels, it can be solely determined by the number of classes.
To focus on the general framework, we assume $p(Y \vert Z)$ is known or estimated beforehand.
Figure~\ref{fig:conditional_probability} illustrates several examples of conditional probability $p(Y \vert Z)$.

\section{METHOD}
\label{sec:method}
In the fully-supervised scenario, where an i.i.d.~sample of $(X, Z)$-pairs $\{x_i, z_i\}_{i=1}^n \iid p(X, Z)$ is given, we can simply estimate the conditional probability $p(Z \vert X)$ from the sample by fitting a discriminative model using the \emph{maximum likelihood}. 
However, it is not the case for learning from indirect observations because $Z$ can not be observed. 
In this section, we propose a general method to handle indirect observations by slightly modifying the maximum likelihood estimator.

Concretely, to predict $Z$ from $X$, we model the conditional probability $p(Z \vert X)$ using a certain parametric distribution, 
e.g., a categorical distribution for the classification problem, or a Gaussian distribution for the regression problem.
The distribution is parameterized by $\theta \in \Theta$, where $\Theta$ denotes the parameter space.
The parameter $\theta$ is determined by $X$ via a deterministic function $f$ parameterized by $W$, such as a deep neural network. i.e.,
\begin{equation}
  p(Z \vert X) 
= 
  p(Z \vert \theta=f(X; W)) 
.
\end{equation}
At this stage, the only content is $p(Z \vert X)$, which is determined by the type of distribution, and the family of deterministic function $f$. 
Differentiability w.r.t.~$W$ is required if we use a gradient method for optimization.

To model the indirect observation $Y$, the key idea is to relate $Y$ with $Z$ using $p(Y \vert Z)$.
Indeed, we can marginalize $p(Y, Z \vert X) = p(Y \vert Z) p(Z \vert X)$ over $Z$ to get the conditional probability $p(Y \vert X)$:
\begin{equation}
\begin{aligned}
\label{eq:p_y_x}
  p(Y \vert X)
&= 
  \int_\sZ
  p(Y \vert z) p(z \vert \theta=f(X; W)) 
  \D{z}
\\
&=
  \E_{Z \sim p(Z \vert \theta=f(X; W))}
  [p(Y \vert Z)]
,
\end{aligned}
\end{equation}
where $\E[\cdot]$ denotes the expectation.

This means that the discriminative model $p(Z \vert \theta=f(X; W))$ for the true target $Z$ is used as a submodule for modeling $P(Y \vert X)$ using $p(Y \vert Z)$.
Note that $p(Y \vert X)$ remains differentiable w.r.t.~$W$.
In this way, we can still use the maximum likelihood method to estimate $p(Y \vert X)$ using $(X, Y)$-pairs without direct observations of $Z$, which leads to an estimator of $p(Z \vert X)$ implicitly.

Concretely, our learning objective, the \emph{expected log-likelihood} given indirect observations, is defined as
\begin{equation}
\label{eq:expected_log_likelihood}
\begin{aligned}
  L(W)
&=
  \E_{X, Y \sim p(X, Y)}
  [\log p(Y \vert X)]
\\
&=
  \E_{X, Y \sim p(X, Y)}
  \brackets*{
  \log
  \E_{Z \sim p(Z \vert \theta=f(X; W))}
  [p(Y \vert Z)]
  }
,
\end{aligned}
\end{equation}
which measures how likely observed data can be generated using a certain parameter $W$ of our model.
Given an i.i.d.~sample of $(X, Y)$-pairs 
$\{x_i, y_i\}_{i=1}^n \iid p(X, Y)$, 
the \emph{empirical distribution} is defined as 
$
  \widehat{p}(X, Y)
= 
  \frac1n \sum_{i=1}^n
  \delta(X - x_i) \delta(Y - y_i)
$,
where $\delta(\cdot)$ denotes the \emph{Dirac delta function}.
It follows from the law of large numbers that
for any measurable real-valued function $f:\sX \times \sY \to \R$ where $\E_{p(X, Y)}[f]$ exists,
$
  \E_{\widehat{p}(X, Y)}[f]
\convergeas 
  \E_{p(X, Y)}[f]
$ 
as $n \to \infty$
\citep[][p.269]{van2000asymptotic}.
Then, let $f = \log p(Y \vert X)$,
we can approximate the expectation in Equation~\eqref{eq:expected_log_likelihood} by the sample mean.
The \emph{log-likelihood} is defined as
\begin{equation}
\begin{aligned}
  \widehat{L}(W)
&=
  \E_{X, Y \sim \widehat{p}(X, Y)}
  [\log p(Y \vert X)]
\\
&=
  \frac1n
  \sum_{i=1}^n
  \brackets*{
  \log
  \E_{Z \sim p(Z \vert \theta = f(x_i; W))}
  [p(y_i \vert Z)]
  }
.
\end{aligned}
\end{equation}
Then, $\widehat{L}(W) \convergeas L(W)$ as $n \to \infty$.
The \emph{maximum likelihood estimator} (MLE) of $W$ given indirect observations is $W^* = \argmax_W \widehat{L}(W)$.
We will analyze when this estimator provides reasonable solution theoretically (Section~\ref{sec:analyses}) and experimentally (Section~\ref{sec:experiments}).

\paragraph{Motivations}
Note that we do not model $p(Y \vert X)$ first and then use that to predict $p(Z \vert X)$ for three reasons.
First, under our assumptions on the data generating process, to get $p(Z \vert X)$ from $p(Y \vert X)$, we need to marginalize $p(Y, Z \vert X) = p(Z \vert X, Y) p(Y \vert X)$ over $Y$. 
However, $p(Z \vert X, Y)$ is not a constant regarding $X$ and is hard to estimate.
Second, according to the \emph{data processing inequality}, the mutual information between $X$ and $Y$ cannot be greater than the mutual information between $X$ and $Z$ \citep{mackay2003information}.
Thus $p(Y \vert X)$ cannot be easier to estimate than $p(Z \vert X)$.
Third, for a certain task, there is only one true target, but there could be many types of indirect observation.
By modeling $p(Z \vert X)$ first and then use it to model different indirect observations $p(Y_1 \vert X), p(Y_2 \vert X), \dots$, we can utilize various sources of weak supervision.
We also demonstrate this experimentally in Section~\ref{ssec:coarse_grained_labels}.

\section{ANALYSES}
\label{sec:analyses}
Although our proposed method in Section~\ref{sec:method} is simple, several fundamental questions remain unanswered.
The first question is whether our method will find the same solution as learning from direct observations.
The second question is how much information we can obtain from a certain type of indirect observation.
In this section, we discuss the \emph{consistency} of the maximum likelihood estimator (MLE) given indirect observations, and then move on to the discussion on its \emph{asymptotic variance}.

Here, we consider a fixed feature vector $x_0 \in \sX$ and consequently a fixed but unknown parameter $\theta_0 = f(x_0; W)$ of the parametric distribution $p(Z \vert \theta_0)$.
Note that different values of $W$ may lead to the same $\theta_0$, i.e., $W$ may not be identifiable.
For example, if we use a deep neural network with a softmax as the last layer for $f$, then $f$ is overparameterized and we can only obtain an observationally equivalent estimator of $W$.
Thus, we analyze the estimation of $\theta_0$ in this section.

\subsection{Consistency: 
Feasibility of Learning from Indirect Observations}
\label{ssec:consistency}
In order to ensure learning from indirect observations is feasible, we need to find conditions when the estimator is \emph{consistent}.
We say that an estimator $\widehat{\theta}_n$ of $\theta$ based on $n$ sample points is consistent if $\widehat{\theta}_n$ converges to the true parameter $\theta_0$ in probability as $n \to \infty$.
Given \emph{direct observation} $Z$, it is well-known that the MLE of $\theta$ exists and is consistent under mild conditions \citep[see e.g.,][]{van2000asymptotic}:
\begin{theorem} 
\label{thm:consistency_direct}
For learning from direct observations $p(Z \vert \theta)$, the MLE of $\theta$ is consistent, if following conditions hold:
\begin{enumerate}[itemsep=-1mm, label=(\Alph*)]
\item identifiability:\\
$
  \forall \theta_1, \theta_2 \in \Theta
, 
  \theta_1 \neq \theta_2 
\implies 
  p(Z \vert \theta_1) \neq p(Z \vert \theta_2)
$
a.e.;
\item compactness:
the parameter space $\Theta$ is compact;
\item differentiability:
$\log p(Z \vert \theta) \in C^1$
w.r.t. $\theta$;
\item i.i.d. observations:
$\{z_i\}_{i=1}^n \iid p(Z \vert \theta_0)$.
\end{enumerate}
\end{theorem}
Here, conditions (B), (C) can be replaced with slightly weaker conditions \citep[see e.g.,][]{van2000asymptotic, lehmann2006theory}.
Nonetheless, condition (A) is necessary for any estimator to be consistent.

Now, consider the indirect observation $Y$.
We need $p(Y \vert \theta)$ to satisfy the above conditions (A)-(D) as well:
(B) remains the same;
(C) can be employed by differentiating Equation~\eqref{eq:p_y_x} under the integral sign;
and (D) is an assumption in our problem setting.
The only nontrivial additional condition is (A),
as follows:
\begin{theorem}
\label{thm:consistency_indirect}
For learning from indirect observations $p(Y \vert \theta)$, compared with learning from direct observations, the only additional requirement for the consistency of the MLE of $\theta$ is the identifiability of $\theta$.

$
  \forall \theta_1, \theta_2 \in \Theta,
  \theta_1 \neq \theta_2 
\implies 
  p(Y \vert \theta_1) \neq p(Y \vert \theta_2)
$
a.e.,

where
$
  p(Y \vert \theta_1) 
= 
  \E_{Z \sim p(Z \vert \theta_1)}
  [p(Y \vert Z)] 
$,
and
$
  p(Y \vert \theta_2)
=
  \E_{Z \sim p(Z \vert \theta_2)}
  [p(Y \vert Z)]
$.
\end{theorem}

\paragraph{Related Work} 
\citet{patrini2017making} proved that in the context of learning from noisy labels (where $\abs{Z} = \abs{Y}$), minimizing a forward correction objective yields the same minimizer as the fully-supervised risk.
Their proof is based on a notion of the proper composite loss~\citep{reid2010composite} and they need to assume the noise transition matrix to be invertible (Theorem 2 of \citet{patrini2017making}).
However, using the inverse of a stochastic matrix $p(Y \vert Z)$ may cause potential problems because elements of the inverse are not necessarily non-negative, unless $p(Y \vert Z)$ is a permutation matrix.
This may lead to a negative estimation of the non-negative risk.
Here, our result interprets forward correction from the maximum likelihood perspective without resorting to the reverse of a stochastic matrix.
For learning from noisy labels, the identifiability of $\theta$ implies the invertibility of the noise transition matrix.
For other types of indirect observation, it can be viewed as a generalization of \citet{patrini2017making} to a scenario where $\abs{Y}$ is not necessarily equal to $\abs{Z}$.

\subsection{Asymptotic Variance: 
Information from Indirect Observations}
\label{ssec:asymptotic_variance}
Even if it is feasible to learn from two types of indirect observation, one could carry more information about the true target and is easier to learn from than the other.
Here, we develop tools for measuring how much information can be obtained from a certain type of indirect observation.

\paragraph{Preliminaries}
First we introduce a few necessary concepts.
Assume $\theta$ is a $K$-dimensional vector.
The \emph{score} function is defined as the gradient of the log-likelihood function w.r.t.~the parameter $\theta$:
\begin{equation}
\begin{aligned}
\label{eq:score}
&
  s(z, \theta)
=
  \diffp{}{\theta} \log p(z \vert \theta)
,
\\
&
  s(y, \theta)
=
  \diffp{}{\theta} \log p(y \vert \theta)
.
\end{aligned}
\end{equation}
The \emph{Fisher information} (in its matrix form) is defined as the variance-covariance matrix of the score function:
\begin{equation}
\label{eq:fisher}
\begin{aligned}
&
  \Info_Z(\theta)
=
  \E_{Z \sim p(Z \vert \theta)}
  \brackets*{s(Z, \theta) s(Z, \theta)\T}
,
\\
&
  \Info_Y(\theta)
=
  \E_{Y \sim p(Y \vert \theta)}
  \brackets*{s(Y, \theta) s(Y, \theta)\T}
.
\end{aligned}
\end{equation}
We emphasize that, in our problem setting, there exist two kinds of Fisher information regarding the same parameter $\theta$, depending on whether the observation is $Z$ or $Y$.
We denote the difference by the subscription.

The Fisher information plays an important role in asymptotic theory \citep{lehmann2006theory}.
For example, the \emph{Cram\'er-Rao bound} provides a lower bound on the variance of any unbiased estimator $\widehat\theta_n$, not necessarily an MLE, in terms of the Fisher information:
$\Cov(\widehat\theta_n) \succeq \Info(\theta)\inv$,
where $\succeq$ indicates the Loewner order\footnote{
Loewner order $\succeq$: let $A$ and $B$ be symmetric matrices. $A \succeq B$ if $A - B$ is positive semi-definite.
}.
Further, if $\widehat\theta_n$ is the MLE under our assumptions in Theorem~\ref{thm:consistency_direct},
then $\widehat\theta_n$ is \emph{asymptotically normal}:
$
  \sqrt{n}(\widehat\theta_n - \theta_0) 
\converged 
  \Normal(0, \Info(\theta)\inv)
$.

We can use the \emph{asymptotic variance}
$[\Info_Y(\theta)\inv]_{i, i}$, 
$(i=1, \dots, K)$\footnote{
This denotes diagonal elements of the inverse of the Fisher information matrix.}
to measure how much information can be obtained from a certain type of indirect observation.
We provide an example for the case where $Z$ and $Y$ are both discrete.


\begin{example}
\label{exm:categorical_distribution}
Consider a special case where both the true target $Z$ and the indirect observation $Y$ follow categorical distributions.
Let the number of classes be
$\abs{Z} = K_Z$, $\abs{Y} = K_Y$, respectively.

Let $p(Z = i) = \theta_i$ and $\theta \in \simplex{K_Z - 1}$, where $\simplex{}$ denotes the standard simplex.
Then, the likelihood is
\begin{equation}
  p(z \vert \theta)
=
  \prod_{i=1}^{K_Z}
  \theta_i^{[z = i]}
=
  \exp \braces*{\sum_{i=1}^{K_Z} [z = i] \log \theta_i}
,
\end{equation}
where $[\cdot]$ denotes the Iverson bracket\footnote{
Iverson bracket $[\cdot]$: $[P] = 1$ if $P$ is true, otherwise $0$. 
}.

The score and the Fisher information regarding the true target $Z$ are
\begin{subequations}
\begin{equation}
  [s(z, \theta)]_i 
= 
  \frac{[z = i]}{\theta_i}
,
\end{equation}
\begin{equation}
  \Info_Z(\theta)
= 
  \diag\braces*{
  \frac1{\theta_1}, \dots, \frac1{\theta_{K_Z}}
  }
.
\end{equation}
\end{subequations}
The asymptotic variance of $\theta_i$ is 
$[\Info_Z(\theta)\inv]_{i, i} = \theta_i$.

Now consider $Y$.
Let $p(Y = j) = \varphi_j$ and $\varphi \in \simplex{K_Y - 1}$,
where
$
  \varphi_j 
= 
  \sum_{i=1}^{K_Z} p(Y = j \vert Z = i) \theta_i
$.

Then, the likelihood is
\begin{equation}
  p(y \vert \theta)
=
  \prod_{j=1}^{K_Y}
  \varphi_j^{[y = j]}
=
  \exp \braces*{\sum_{j=1}^{K_Y} [y = j] \log \varphi_j}
,
\end{equation}

The score and the Fisher information regarding the indirect observation $Y$ are
\begin{subequations}
\begin{equation}
  [s(y, \theta)]_i 
= 
  \sum_{j=1}^{K_Y}
  \frac{[y = j] p(Y = j \vert Z = i)}{\varphi_j}
,
\end{equation}
\begin{equation}
  [\Info_Y(\theta)]_{i_1, i_2}
= 
  \sum_{j=1}^{K_Y}
  \frac
  {p(Y = j \vert Z = i_1) p(Y = j \vert Z = i_2)}
  {\varphi_j}
.
\end{equation}
\end{subequations}
It is not easy to compute the inverse of this Fisher information matrix.
However, the reciprocal of diagonal elements gives
\begin{equation}
\begin{aligned}
  [\Info_Y(\theta)]_{i, i}\inv 
&= 
  \brackets*{
  \sum_{j=1}^{K_Y} p(Y = j \vert Z = i)p(Z = i \vert Y = j)
  }\inv
  \theta_i
\\
&\geq
  \brackets*{
  \sum_{j=1}^{K_Y} p(Y = j \vert Z = i)
  }\inv
  \theta_i
=
  \theta_i
.
\end{aligned}
\end{equation}
Because 
$
  [A\inv]_{i, i} 
\geq
  [A]_{i, i}\inv
$
holds for any positive definite matrix $A$,
we have
\begin{equation}
\label{eq:fisher_information_inequality}
  [\Info_Y(\theta)\inv]_{i, i} 
\geq
  [\Info_Y(\theta)]_{i, i}\inv
\geq
  [\Info_Z(\theta)\inv]_{i, i}
.
\end{equation}

\end{example}


We can generalize Inequality~\eqref{eq:fisher_information_inequality} in Example~\ref{exm:categorical_distribution}, and show that learning from indirect observations cannot be as statistically efficient as learning from direct observations, as stated in Theorem~\ref{thm:fisher_information_inequality}.
We defer its proof to Appendix~\ref{app:fisher_information_inequality}.

\begin{theorem}
\label{thm:fisher_information_inequality}
$\Info_Y(\theta)\inv \succeq \Info_Z(\theta)\inv$.
i.e., the asymptotic variance of the MLE based on indirect observations is always not less than the one based on direct observations.
\end{theorem}

Nonetheless, analyzing the asymptotic variance provides a tool to balance the trade-off between the quality of labels and costs of the label collection process.
If the asymptotic variance is large, we might need a relatively large number of data points to acquire sufficient predictive power.
For example, if a certain weak supervision costs $\frac{1}{10}$ of costs of the true target, but its asymptotic variance is $100$ times larger, it might be more reasonable to collect true labels or find other kinds of weak supervision.

\section{EXPERIMENTS}
\label{sec:experiments}
In this section, we propose two novel problem settings that are examples of learning from indirect observations, and conduct experiments to show practical usefulness of our framework.

\begin{figure*}[t]
\centering
\begin{subfigure}[t]{.3\textwidth}
  \centering
  \includegraphics[width=\linewidth]
  {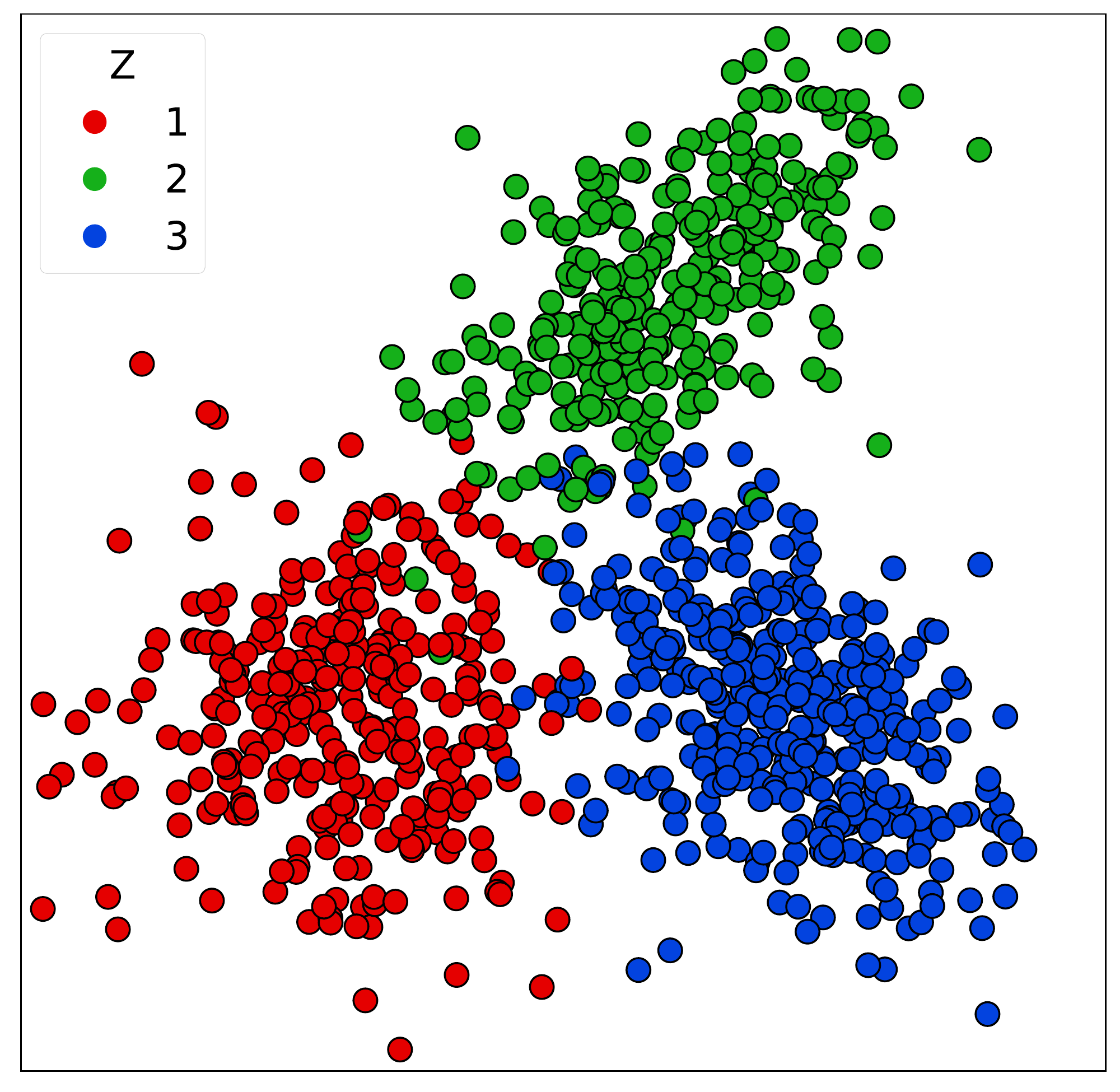}
  \caption{direct observation Z}
  \label{sfig:direct_observation_Z}
\end{subfigure}%
\hfill
\begin{subfigure}[t]{.3\textwidth}
  \centering
  \includegraphics[width=\linewidth]
  {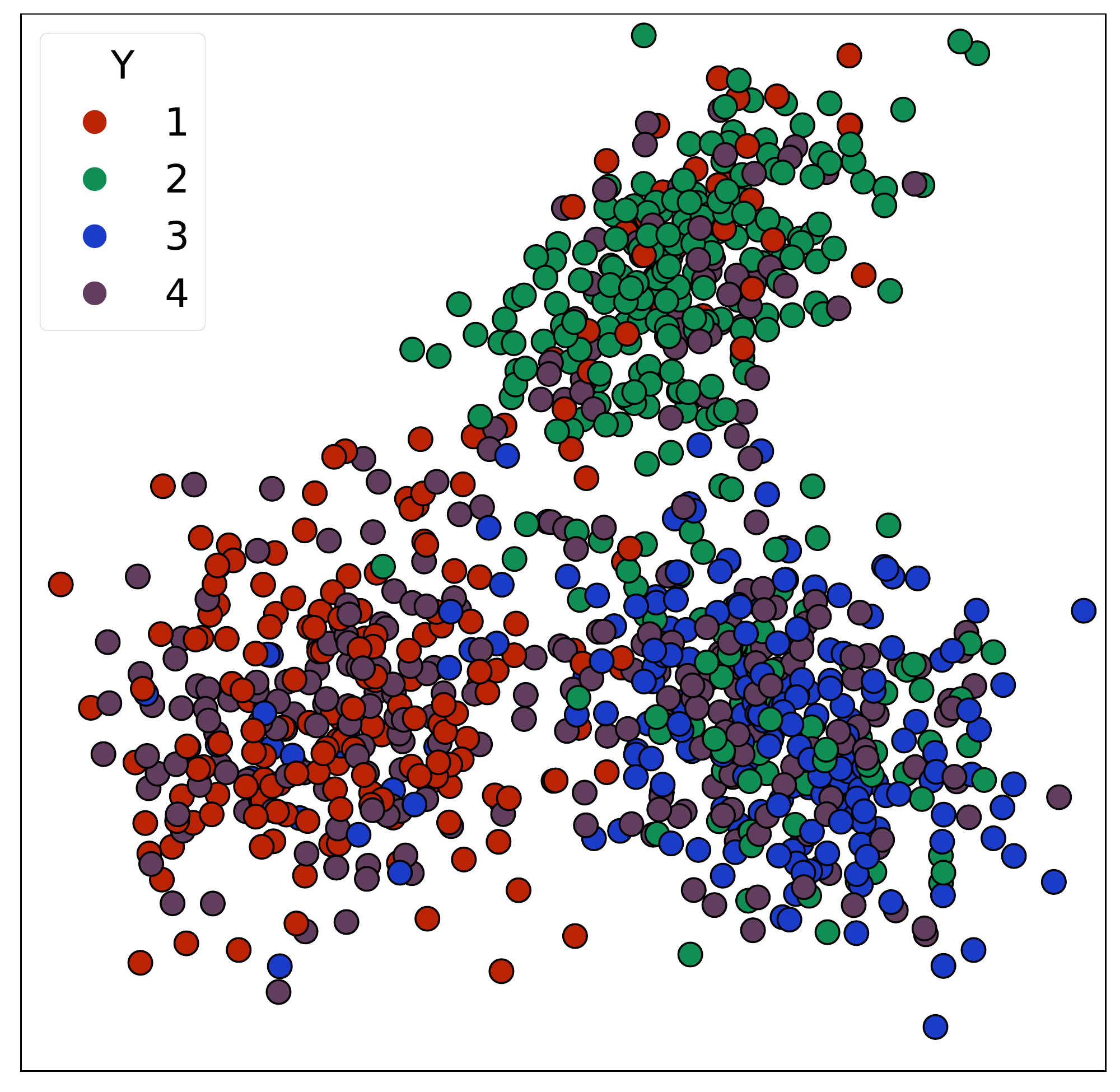}
  \caption{indirect observation Y}
  \label{sfig:indirect_observation_Y}
\end{subfigure}%
\hfill
\begin{subfigure}[t]{.3\textwidth}
  \centering
  \includegraphics[width=\linewidth]
  {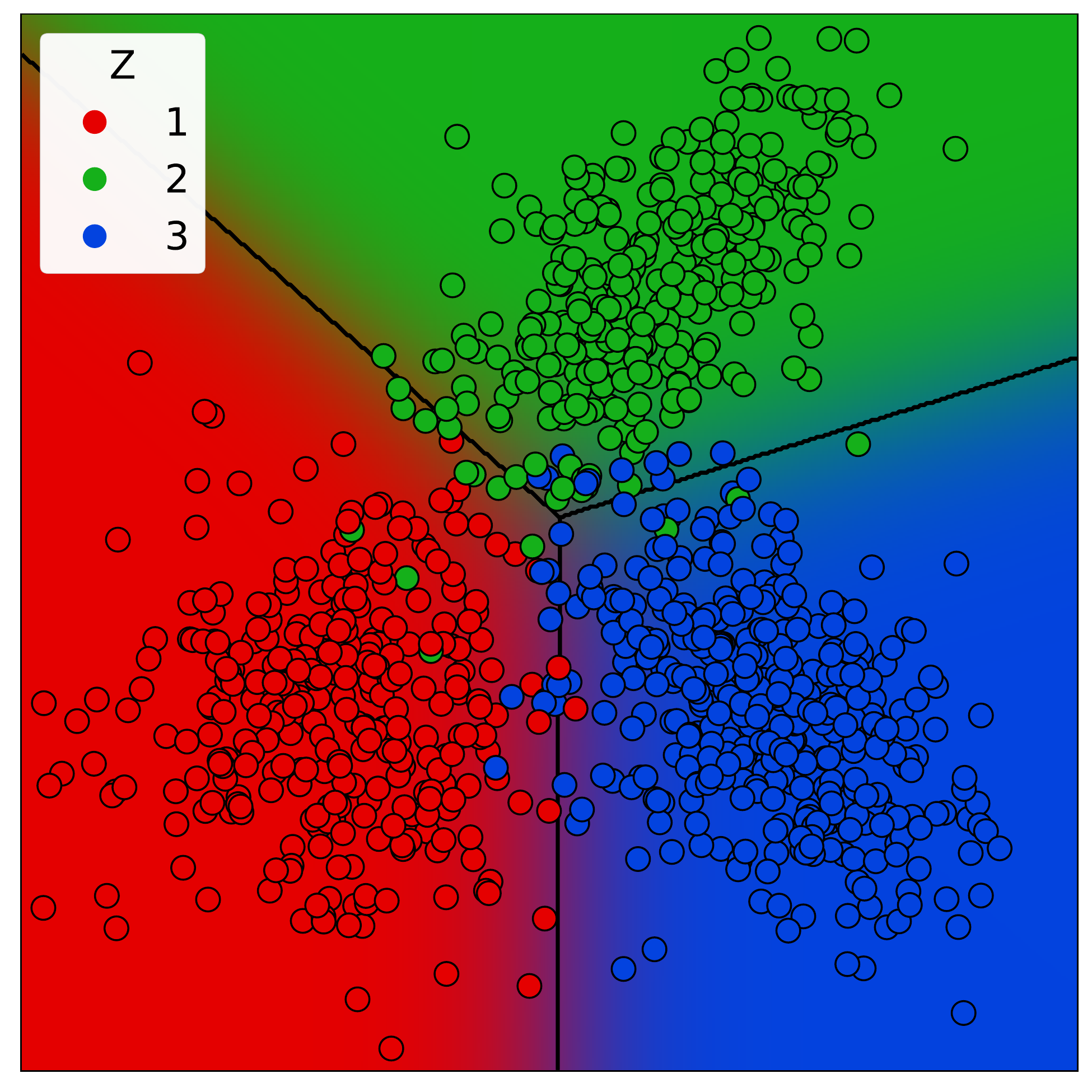}
  \caption{classification result}
    \label{sfig:result}
\end{subfigure}%
\caption{
  \textbf{Learning from multiclass label proportions on synthetic data.}
  (\ref{sfig:direct_observation_Z}) 
  Random sample of the true target $Z$. 
  $3$ classes are marked by color R, G, B, respectively;
  (\ref{sfig:indirect_observation_Y}) 
  Random sample of the indirect observation $Y$. 
  $4$ groups are marked by RGB interpolation using corresponding label proportions;
  (\ref{sfig:result}) 
  Classification result using sample in Figure~\ref{sfig:indirect_observation_Y}, illustrated by RGB interpolation using the predicted probability.
  Sample in Figure~\ref{sfig:direct_observation_Z} and decision boundaries are also plotted.
  Note that the marginal distribution of $X$ (ignoring colors in \ref{sfig:direct_observation_Z} and \ref{sfig:indirect_observation_Y}) should match the true distribution.
}
\label{fig:label_proportions_synthetic_data}
\end{figure*}
\begin{table*}[t]
\caption{\textbf{Accuracy of learning from label proportions on Adult dataset.} 
Means and standard deviations in percentage for 10 trials are reported.
The baseline is learning from direct observations (rightmost column).}
\label{tab:adult}
\begin{center}
\begin{tabular}{lcccc}
\toprule
target (\#classes)
& \multicolumn{3}{c}{
\begin{tabular}[c]{@{}c@{}}
label proportions observation\\ 
grouping attribute (\#groups)
\end{tabular}} 
& direct observation
\\ \cmidrule{2-4}
& \texttt{education} (8) 
& \texttt{occupation} (14) 
& \texttt{relationship} (6) 
&  
\\ \midrule
\texttt{income} (2) 
& $76.73 \pm 1.77$ 
& $78.02 \pm 1.09$
& $77.60 \pm 1.02$
& $80.42 \pm 0.28$
\\
\texttt{marital-status} (3) 
& $49.10 \pm 2.41$
& $56.62 \pm 1.03$
& $67.90 \pm 1.89$
& $70.68 \pm 0.08$
\\ \bottomrule
\end{tabular}
\end{center}
\end{table*}
\begin{table*}[t]
\caption{\textbf{Accuracy of learning from coarse-grained/complementary labels on CIFAR-10 dataset.} 
Means and standard deviations in percentage for 5 trials are reported.}
\label{tab:coarse}
\begin{center}
\begin{threeparttable}
\begin{tabular}{ccc}
\toprule
60000 coarse-grained labels & 
60000 complementary labels & 
60000 ordinary labels\tnote{*}
\\
\midrule
$38.72 \pm 1.10$ & 
$67.25 \pm 0.89$ & 
$93.10 \pm 0.18$
\\
\toprule
\begin{tabular}[t]{@{}l@{}}
  $60000$ coarse-grained labels + \\ 
  $10000$ ordinary labels\tnote{*}
\end{tabular} & 
\begin{tabular}[t]{@{}l@{}}
  $60000$ complementary labels + \\ 
  $10000$ ordinary labels\tnote{*}
\end{tabular} & 
\begin{tabular}[t]{@{}l@{}}
  $60000$ coarse-grained labels + \\ 
  $60000$ complementary labels
\end{tabular}
\\
\midrule
$90.05 \pm 0.29$ & 
$85.70 \pm 0.34$ & 
$88.43 \pm 0.23$
\\
\bottomrule
\end{tabular}
\begin{tablenotes}
\footnotesize
\item[*] the true target $Z$
\end{tablenotes}
\end{threeparttable}
\end{center}
\end{table*}

\subsection{Learning from Label Proportions}
\label{ssec:label_proportions}
\emph{Learning from label proportions} (LLP) has been studied in \citet{kuck2012learning, quadrianto2009estimating, felix2013psvm, patrini2014almost, yu2014learning}, but prior studies only focused on the binary case.
In this setting, instead of the label of each instance, only proportions of positive sample points in a group (also called a ``\emph{bag}'') can be observed.
Previous approaches either only work on binary classification, e.g., a support vector machine based method \citep{felix2013psvm}, or only work with a linear classifier \citep{patrini2014almost}.
To deal with multiclass classification, existing methods (e.g., \citet{patrini2014almost}) usually resort to \emph{one-against-all} transformation to binary classification.

Assuming instances are conditionally independent given the group, we can naturally extend LLP to the multiclass case in our framework.
Concretely, let $Z$ and $Y$ be categorically distributed random variables representing classes and groups (``bags''), respectively.
In this setting, we can obtain information about proportions of each class in each group and use it as an estimator of $P(Z \vert Y)$.
Then, $P(Y)$ can be estimated by the frequency in the dataset and $P(Y \vert Z)$ can be calculated via Bayes' rule.
If $P(Y \vert Z, X) = P(Y \vert Z)$ holds, then we can apply the maximum likelihood method described in Section~\ref{sec:method}, i.e., just estimate the probability of groups $Y$, and subsequently get predictions of classes $Z$.


\paragraph{Synthetic Dataset}
First we constructed a synthetic dataset (Figure~\ref{fig:label_proportions_synthetic_data}) to showcase the problem setting of multiclass LLP and the feasibility of the maximum likelihood method.
Consider two-dimensional feature vectors $X$ which can be classified into $\abs{Z} = 3$ classes.
$\abs{Y} = 4$ groups of data are collected, whose label proportions can be observed.
The visualization of data and the classification result are shown in Figure~\ref{fig:label_proportions_synthetic_data}.
Experiment details can be found in Appendix~\ref{app:experiment_details}.

We can see that $3$ classes can be classified using $4$ groups of observations where only label proportions in each group can be observed (Figure~\ref{sfig:result}).
This experiment also illustrates the limitation of our method. 
First, i.i.d.~observations of $Y$ are required, i.e., the marginal distribution of $X$ should match the true distribution (Figure~\ref{sfig:direct_observation_Z} and Figure~\ref{sfig:indirect_observation_Y}).
This assumption usually holds when the data is collected altogether and overrepresentation and underrepresentation are avoided by design, such as vote data and census data.
But it may be violated when data from each group are collected separately.
Second, $P(Y \vert Z, X) = P(Y \vert Z)$ should hold.
We will also show the influence of this assumption in the next experiment.


\paragraph{Adult Dataset}
We further demonstrate the feasibility of multiclass LLP on real-world data,
and show how the result depends on the assumption 
$P(Y \vert Z, X) = P(Y \vert Z)$.
We use the UCI Adult dataset\footnote{
UCI Machine Learning Repository, Adult dataset
\citep{UCI}\\
\url{http://archive.ics.uci.edu/ml/datasets/Adult}
},
a subset of 1994 census database.
The original task is to predict whether a person makes over $50K$ a year based on some demographic factors, such as age, sex, education and occupation.
This dataset has been adapted to verify algorithms for LLP \citep{yu2014learning, patrini2014almost}.

Here, we consider a binary attribute \texttt{income} and a multiclass attribute \texttt{marital-status} privacy-sensitive and thus not revealed, but whose proportions in some demographic groups can be estimated.
To better imitate the data collection process in real-world scenarios, we follow procedures used in \citet{yu2014learning, patrini2014almost}.
First we split the dataset into groups $Y$ according to a selected attribute (e.g., \texttt{education}), calculated the frequency of the true target $Z$ (e.g., \texttt{income}) in each group, and then removed the true target $Z$ from the data.
We want to use other attributes $X$ (\texttt{age}, \texttt{sex}, \texttt{hours-per-week}, etc.) to predict $Z$ given only groups $Y$.
Preprocessing procedures and experiment details can be found in Appendix~\ref{app:experiment_details}.

The results are listed in Table~\ref{tab:adult}.
We can observe that knowing \texttt{education}, \texttt{occupation} or \texttt{relationship}, and proportions of high-income people in each group, we can train a classifier that is comparable with the one trained from direct observations of \texttt{income}.
The accuracy gap can be less than $5\%$.
For the multiclass attribute \texttt{marital-status}, because \texttt{relationship} carries almost the same amount of information about \texttt{marital-status}, i.e., $P(Y \vert Z, X) = P(Y \vert Z)$, the accuracy gap is only around $2\%$.
Meanwhile \texttt{education} and \texttt{occupation} provide relatively lower predictive power than \texttt{relationship} for \texttt{marital-status}.
This illustrates that if $P(Y \vert Z, X) = P(Y \vert Z)$ holds, our method works relatively well on real-world data.

\subsection{Learning from Coarse-grained Labels}
\label{ssec:coarse_grained_labels}
Next, we study a novel problem setting called \emph{learning from coarse-grained labels}.
Previous studies on multiclass classification usually assume concepts of labels to be mutually exclusive and at the same granularity level.
However, labels often have a hierarchical structure in real-world problems (e.g., CIFAR-100 \citep{CIFAR}, ImageNet \citep{ImageNet}).
Sometimes, we can only obtain coarse-grained level annotations about the instance, namely \emph{coarse-grained labels}, e.g., genus level annotation of an animal image instead of species or breed level annotation.

This problem arises naturally, e.g., (1) when we want to collect data from the internet using a web scraper, and we do not want to waste some labeled data that is not as fine-grained as we want,
or (2) when we want to refine a classifier, but newly collected fine-grained labels are scarce while existing coarse-grained labels are abundant.

This problem setting can be interpreted as an example of learning from indirect observations.
Concretely, let $Z$ and $Y$ be random variables representing the \emph{fine-grained label} and the \emph{coarse-grained label}, respectively. 
The set of fine-grained labels is partitioned into a number of disjoint (non-overlapping) subsets as coarse-grained labels.
Thus, the conditional probability $P(Y \vert Z)$ is solely determined by the hierarchy of labels.
An example is illustrated in Figure~\ref{fig:conditional_probability}.
Then, we can apply the maximum likelihood method described in Section~\ref{sec:method} to utilize coarse-grained labels.

However, it is notable that according to Theorem~\ref{thm:consistency_indirect}, it is \emph{impossible} to learn from only coarse-grained labels because the parameter $\theta$ is not identifiable in this scenario.
The model cannot distinguish fine-grained labels in a coarse-grained label group \emph{without any regularization}, e.g., regularization on the marginal distribution $p(Z)$, or manifold regularization.
Thus, we focused on the scenario where a small number of fine-grained observations or other kinds of weak supervision are available.

We evaluated our method on the CIFAR-10 dataset\footnote{
The CIFAR-10 dataset
\citep{CIFAR}\\
\url{https://www.cs.toronto.edu/~kriz/cifar.html}
},
which consists of $60000$ $32\times32$ colour images in $10$ classes:
(\texttt{airplane}, \texttt{ship}), 
(\texttt{automobile}, \texttt{trunk}),
(\texttt{bird}, \texttt{deer}, \texttt{frog}),
(\texttt{horse}, \texttt{cat}, \texttt{dog}).
We can group fine-grained labels into coarse-grained labels by their semantic meanings as parenthesized above:
\texttt{large-vehicle}, \texttt{small-vehicle}, \texttt{wild-animal}, and \texttt{domestic-animal}.
For direct observations, a random sample of size $10000$ was extracted from the original training set.
For indirect observations, we considered coarse-grained labels and also complementary labels.
We used a ResNet-18 \citep{he2016deep, he2016identity} for $p(Z \vert X)$, and used Stochastic Gradient Descent (SGD) with momentum \citep{sutskever2013importance} to optimize the model.
Hyperparameters, training procedures, and other experiment details can be found in Appendix~\ref{app:experiment_details}.

The results are listed in Table~\ref{tab:coarse}.
We can observe that learning from only coarse-grained labels is infeasible, but with a small number of fine-grained labels, it can achieve relatively high accuracy that is comparable with learning from a large number of fine-grained labels.
Integrating different sources of weak supervision such as coarse-grained labels and complementary labels also achieved promising performance.

\section{CONCLUSIONS}
We have introduced a general framework for weakly-supervised learning, \emph{learning from indirect observations}, that includes several existing problems and can give rise to new settings.
We proposed a method based on the \emph{maximum likelihood} principle, which can be straightforwardly implemented for deep neural networks and combine different kinds of weak supervision.
We proposed two novel problem settings under this framework:
\emph{learning from multiclass label proportions}, and \emph{learning from coarse-grained labels}.
The feasibility and advantages of our method are reflected in experimental results.

\subsubsection*{ACKNOWLEDGMENTS}
We thank Ikko Yamane for helpful discussion. 
NC was supported by MEXT scholarship and JST AIP Challenge. 
MS was supported by JST CREST Grant Number JPMJCR18A2.

\subsubsection*{REFERENCES}
\bibliography{references}

\newpage
\onecolumn

\appendix
\part*{\Large{Appendix}}

\section{PU Learning: Censoring Setting \& Case-control Setting}
\label{app:pu}
In this section, we illustrate the difference between the \emph{censoring setting} \citep{elkan2008learning} and the \emph{case-control setting} \citep{ward2009presence, du2014analysis} of PU learning.
The same issue arises in other scenarios, e.g., the \emph{class-conditional noise} model \citep{angluin1988learning, natarajan2013learning, patrini2017making} and the \emph{mutual contamination} model \citep{scott2013classification, blanchard2014decontamination, menon2015learning} for learning from noisy labels.
The difference between those two settings shows what kind of problem our framework can cover, and what kind of problem can not be solved using our method.

In the censoring setting, the whole dataset is collected first and then a small number of positive sample points are picked out randomly (label censoring procedure).
Thus it is a special case of learning from indirect observations with $\abs{Z} = \abs{Y} = 2$ and 
$p(Y = U \vert Z = N) = 1$.
In the case-control setting, the positive sample and the unlabeled sample are drawn separately. The unlabeled sample is regarded as drawn from the marginal distribution.

For example, assuming the positive : negative ratio is $1:1$, an example of the number of data points in each class in two settings is shown in Table~\ref{tab:censoring} and Table~\ref{tab:case_control}, respectively.
In the censoring setting, $100$ data points are collected altogether and $20$ positive data points are picked out, leaving $30$ positive  and all $50$ negative data points unlabeled;
In the case-control setting, $20$ positive and $80$ unlabeled data points are collected separately.
There are $40$ positive and $40$ negative data points in the unlabeled sample.
If we treat the data collection process incorrectly, we will introduce a selection bias which degenerates the performance of the model.
\begin{table}[!h]
\centering
\begin{minipage}{.3\linewidth}
  \caption{Censoring setting}
  \label{tab:censoring}
  \centering
  \begin{tabular}{|c|cc|c|}
  \hline
    & P    & N    &       \\ \hline
  P & $20$ & $0$  & $20$  \\
  U & $30$ & $50$ & $80$  \\ \hline
    & $50$ & $50$ & $100$ \\ \hline
  \end{tabular}
\end{minipage}%
\begin{minipage}{.3\linewidth}
  \caption{Case-control setting}
  \label{tab:case_control}
  \centering
  \begin{tabular}{|c|cc|c|}
  \hline
    & P    & N    &        \\ \hline
  P & $20$ & $0$   & $20$  \\
  U & $40$ & $40$  & $80$ \\ \hline
    & $60$ & $40$  & $100$ \\ \hline
  \end{tabular}
\end{minipage} 
\end{table}

\section{Proof of the Fisher Information Inequality}
\label{app:fisher_information_inequality}
In this section, we prove the Theorem~\ref{thm:fisher_information_inequality}:
$\Info_Z(\theta) \succeq \Info_Y(\theta)$, 
and therefore 
$\Info_Y(\theta)\inv \succeq \Info_Z(\theta)\inv$.

We assume $\theta$ is a $K$-dimensional vector, 
so $s(z, \theta)$ and $s(y, \theta)$ are also $K$-dimensional vectors, 
while $\Info_Z(\theta)$ and $\Info_Y(\theta)$ are $K \times K$ matrices.

As defined in Equation~\eqref{eq:score}, the score function for the indirect observation $Y$, $s(y, \theta)$, can be written in terms of $p(Y \vert Z)$ and $s(z, \theta)$ as
\begin{equation}
\begin{aligned}
  s(y, \theta)
&=
  \diffp{}{\theta} \log p(y \vert \theta)
=
  \diffp{}{\theta} 
  \log \int_\sZ 
  p(y \vert z) p(z \vert \theta) 
  \D{z}
\\
&=
  \frac
  {\displaystyle
    \E_{Z \sim p(Z \vert \theta)} 
    \brackets*{p(y \vert Z) s(Z, \theta)}
  }
  {\displaystyle
    \E_{Z \sim p(Z \vert \theta)} 
    \brackets*{p(y \vert Z)}
  }
.
\end{aligned}
\end{equation}

As defined in Equation~\eqref{eq:fisher}, the Fisher information matrix $\Info_Z(\theta)$ is defined as
\begin{equation}
  \Info_Z(\theta)
=
  \E_{Z \sim p(Z \vert \theta)}
  \brackets*{s(Z, \theta) s(Z, \theta)\T}
.
\end{equation}

And the Fisher information matrix $\Info_Y(\theta)$ can be written in terms of $p(Y \vert Z)$ and $s(z, \theta)$ as
\begin{equation}
\begin{aligned}
  \Info_Y(\theta)
&=
  \E_{Y \sim p(Y \vert \theta)}
  \brackets*{s(Y, \theta) s(Y, \theta)\T}
\\
&=
  \E_{Y \sim p(Y \vert \theta)}
  \brackets*{
  \frac
  {\displaystyle
    \E_{Z \sim p(Z \vert \theta)} 
    \brackets*{p(y \vert Z) s(Z, \theta)}
    \E_{Z \sim p(Z \vert \theta)} 
    \brackets*{p(y \vert Z) s(Z, \theta)\T}
  }
  {\displaystyle
    \E_{Z \sim p(Z \vert \theta)} 
    \brackets*{p(y \vert Z)}^2
  }
  }
\end{aligned}
\end{equation}

To prove 
$\Info_Z(\theta) \succeq \Info_Y(\theta)$, we need to prove 
$\Info_Z(\theta) - \Info_Y(\theta)$ is a positive semidefinite matrix.
i.e., $\forall t \in \R^K$,
$
  t\T 
  \brackets*{\Info_Z(\theta) - \Info_Y(\theta)} 
  t 
  \geq 0
$.

Let $t$ be any vector in $\R^K$.
By the linearity of the expectation operator, we have
\begin{equation}
  t\T \Info_Z(\theta) t 
=
  t\T 
  \E_{Z \sim p(Z \vert \theta)}
  \brackets*{s(Z, \theta) s(Z, \theta)\T} 
  t
=
  \E_{Z \sim p(Z \vert \theta)}
  \brackets*{t\T s(Z, \theta) s(Z, \theta)\T t}
=
  \E_{Z \sim p(Z \vert \theta)}
  \brackets*{(t\T s(Z, \theta))^2}
,
\end{equation}
and
\begin{equation}
  t\T \Info_Y(\theta) t 
=
  \E_{Y \sim p(Y \vert \theta)}
  \brackets*{
  \frac
  {\displaystyle
    \E_{Z \sim p(Z \vert \theta)} 
    \brackets*{p(y \vert Z) t\T s(Z, \theta)}^2
  }
  {\displaystyle
    \E_{Z \sim p(Z \vert \theta)} 
    \brackets*{p(y \vert Z)}^2
  }
  }
.
\end{equation}

Therefore,
\begin{equation}
\label{eq:tT_Z_Y_t}
\begin{aligned}
&
  t\T 
  \brackets*{\Info_Z(\theta) - \Info_Y(\theta)} 
  t
\\
=&
  \E_{Z \sim p(Z \vert \theta)}
  \brackets*{(t\T s(Z, \theta))^2}
-
  \E_{Y \sim p(Y \vert \theta)}
  \brackets*{
  \frac
  {\displaystyle
    \E_{Z \sim p(Z \vert \theta)} 
    \brackets*{p(y \vert Z) t\T s(Z, \theta)}^2
  }
  {\displaystyle
    \E_{Z \sim p(Z \vert \theta)} 
    \brackets*{p(y \vert Z)}^2
  }
  }
\\
=&
  \E_{Y \sim p(Y \vert \theta)}
  \brackets*{
  \E_{Z \sim p(Z \vert \theta)}
  \brackets*{(t\T s(Z, \theta))^2}
-
  \frac
  {\displaystyle
    \E_{Z \sim p(Z \vert \theta)} 
    \brackets*{p(y \vert Z) t\T s(Z, \theta)}^2
  }
  {\displaystyle
    \E_{Z \sim p(Z \vert \theta)} 
    \brackets*{p(y \vert Z)}^2
  }
  }
\\
=&
  \E_{Y \sim p(Y \vert \theta)}
  \brackets*{
  \frac
  {\displaystyle
    \E_{Z \sim p(Z \vert \theta)}
    \brackets*{p(y \vert Z)}^2
    \E_{Z \sim p(Z \vert \theta)}
    \brackets*{(t\T s(Z, \theta))^2}
  -
    \E_{Z \sim p(Z \vert \theta)} 
    \brackets*{p(y \vert Z) t\T s(Z, \theta)}^2
  }
  {\displaystyle
    \E_{Z \sim p(Z \vert \theta)} 
    \brackets*{p(y \vert Z)}^2
  }
  }
.
\end{aligned}
\end{equation}

The denominator of Equation~\eqref{eq:tT_Z_Y_t} is positive.
We only need to prove that the numerator is non-negative.

By Jensen's inequality, we have
\begin{equation}
  \E_{Z \sim p(Z \vert \theta)}
  \brackets*{(t\T s(Z, \theta))^2}
=
  \E_{Z \sim p(Z \vert \theta)}
  \brackets*{\abs*{t\T s(Z, \theta)}^2}
\geq
  \E_{Z \sim p(Z \vert \theta)}
  \brackets*{\abs*{t\T s(Z, \theta)}}^2
,
\end{equation}
and 
\begin{equation}
  \E_{Z \sim p(Z \vert \theta)} 
  \brackets*{\abs*{p(y \vert Z) t\T s(Z, \theta)}}
\geq
  \abs*{
  \E_{Z \sim p(Z \vert \theta)} 
  \brackets*{p(y \vert Z) t\T s(Z, \theta)}
  }
.
\end{equation}

By H\"{o}lder's inequality, we have
\begin{equation}
  \E_{Z \sim p(Z \vert \theta)}
  \brackets*{p(y \vert Z)}
  \E_{Z \sim p(Z \vert \theta)}
  \brackets*{\abs*{t\T s(Z, \theta)}}
\geq
  \E_{Z \sim p(Z \vert \theta)}
  \brackets*{\abs*{p(y \vert Z) t\T s(Z, \theta)}}
.
\end{equation}

Applying above inequalities, the numerator of Equation~\eqref{eq:tT_Z_Y_t} is
\begin{equation}
\begin{aligned}
&
  \E_{Z \sim p(Z \vert \theta)}
  \brackets*{p(y \vert Z)}^2
  \E_{Z \sim p(Z \vert \theta)}
  \brackets*{(t\T s(Z, \theta))^2}
-
  \E_{Z \sim p(Z \vert \theta)} 
  \brackets*{p(y \vert Z) t\T s(Z, \theta)}^2
\\
\geq&
  \parens*{
  \E_{Z \sim p(Z \vert \theta)}
  \brackets*{p(y \vert Z)}
  \E_{Z \sim p(Z \vert \theta)}
  \brackets*{\abs*{t\T s(Z, \theta)}}
  }^2
-
  \E_{Z \sim p(Z \vert \theta)} 
  \brackets*{p(y \vert Z) t\T s(Z, \theta)}^2
\\
\geq&
  \E_{Z \sim p(Z \vert \theta)}
  \brackets*{\abs*{p(y \vert Z) t\T s(Z, \theta)}}^2
-
  \E_{Z \sim p(Z \vert \theta)} 
  \brackets*{p(y \vert Z) t\T s(Z, \theta)}^2
\\
\geq&
  0
.
\end{aligned}
\end{equation}

Therefore, 
$
  t\T 
  \brackets*{\Info_Z(\theta) - \Info_Y(\theta)} 
  t 
  \geq 0
$
for all $t \in \R^K$.
$\Info_Z(\theta) - \Info_Y(\theta)$ 
is positive semidefinite,
i.e.,
$\Info_Z(\theta) \succeq \Info_Y(\theta)$.

Q.E.D.

\section{Experiment Details}
\label{app:experiment_details}
In this section, we provide missing experiment details in Section~\ref{sec:experiments}.


\subsection{Learning from label proportions on synthetic dataset (Section~\ref{ssec:label_proportions})}
\paragraph{Data}
Feature vectors $X$ were drawn from a Gaussian mixture of $3$ components, while the true target $Z$ is the component indicator.
Indirect observations $Y$ were generated according to a manually defined conditional probability $p(Y \vert Z)$ (a $4 \times 3$ matrix), i.e., strictly according to our assumption in Equation~\ref{eq:factorization}.
$1000$ data points were drawn for the training data and the test data, respectively.

\paragraph{Model}
A linear model was used for $p(Z \vert X)$.
i.e., 
$f(x; W) = \softmax(w\T x + b), \forall x \in \R^2$,
where 
$w \in \R^{3 \times 2}$, 
$b \in \R^{3}$, 
and $W = \{w, b\}$.
The softmax function is applied to get the parameter in the simplex.

\paragraph{Optimization}
We used a Gradient Descent optimizer with a fixed learning rate $0.1$.
The model was trained for total $500$ iterations.


\subsection{Learning from label proportions on Adult dataset (Section~\ref{ssec:label_proportions})}

\paragraph{Data Preprocessing}
There are originally $14 + 1$ attributes:
\texttt{age}, \texttt{workclass}, \texttt{fnlwgt}, \texttt{education}, \texttt{education-num}, \texttt{marital-status}, \texttt{occupation}, \texttt{relationship}, \texttt{race}, \texttt{sex}, \texttt{capital-gain}, \texttt{capital-loss}, \texttt{hours-per-week}, \texttt{native-country}, and \texttt{income}.
Two attributes \texttt{workclass} and \texttt{fnlwgt} were dropped;
Two attributes \texttt{capital-gain} and \texttt{capital-loss} were merged into one attribute \texttt{capital-change} by their difference;
Some classes of four attributes, \texttt{race}, \texttt{education}, \texttt{marital-status}, and \texttt{native-country} were grouped, respectively.

\paragraph{Data}
For all sub-experiments, we used only $7$ attributes for the feature vector $X$: \texttt{age}, \texttt{education-num}, \texttt{race}, \texttt{sex}, \texttt{capital-change}, \texttt{hours-per-week}, and \texttt{native-country}.
The training data were generated as described in Section~\ref{ssec:label_proportions}.

\paragraph{Model}
A linear model was used for $p(Z \vert X)$.

\paragraph{Optimization}
We used an Adam \citep{kingma2014adam} optimizer to train the model. 
The learning rate is initially \num{1e-4} and decays exponentially every epoch with a decaying rate $0.98$. 
$\beta_1 = 0.9$ and $\beta_2 = 0.999$.
The batch size is $128$, and the model is trained for $50$ epochs.


\subsection{Learning from coarse-grained/complementary labels on CIFAR-10 dataset (Section~\ref{ssec:coarse_grained_labels})}

\paragraph{Data}
As stated in Section~\ref{ssec:coarse_grained_labels}, fine-grained labels were grouped into coarse-grained labels by their semantic meanings.
For complementary labels, uniform complementary labels were used.
i.e., $p(Y = j \vert X = i) = 0$ if $i = j$ and $\frac19$ if $i \neq j$.

\paragraph{Model}
We used a modified ResNet-18 \citep{he2016deep, he2016identity} for $p(Z \vert X)$ that takes $32 \times 32$ RGB images as the input.

\paragraph{Optimization}
We used a Stochastic Gradient Descent (SGD) optimizer with momentum \citep{sutskever2013importance} to train the model.
The momentum is $0.9$ and the weight decay ($\ell_2$-regularization) parameter is \num{5e-4}.
The batch size is $128$, and the model is trained for $50$ epochs.
We used a ``warmup-decay'' schedule for the learning rate to accelerate the training.
Concretely, the learning rate increases linearly from $0$ to $0.1$ for $15$ epochs and then decreases exponentially with a decaying rate $0.95$.

\end{document}